\documentclass[runningheads]{llncs}
\usepackage{graphicx}
\usepackage{comment}
\usepackage{amsmath,amssymb} 
\usepackage{xcolor}
\usepackage{tikz}
\usepackage{float}
\usepackage{url}

\usepackage{booktabs} 

\begin{document}
\pagestyle{headings}
\mainmatter
\def\ECCVSubNumber{3220}  

\newcommand{\todo}[1]{\textcolor{red}{[TODO: #1]}}


\setlength{\belowcaptionskip}{-10pt}

\iftrue     
\newcommand{\dimpp}[1]{\textcolor{green}{[DP: #1]}}
\newcommand{\ethan}[1]{\textcolor{blue}{[EW: #1]}}
\newcommand{\antonio}[1]{\textcolor{magenta}{[AT: #1]}}
\newcommand{\ferda}[1]{\textcolor{red}{[FO: #1]}}
\newcommand{\agata}[1]{\textcolor{cyan}{[AL: #1]}}
\newcommand{\imran}[1]{\textcolor{brown}{[IM: #1]}}
\else
\newcommand{\dimpp}[1]{\textcolor{green}{\noindent}}
\newcommand{\ethan}[1]{\textcolor{blue}{\noindent}}
\newcommand{\antonio}[1]{\textcolor{magenta}{\noindent}}
\newcommand{\ferda}[1]{\textcolor{red}{\noindent}}
\newcommand{\agata}[1]{\textcolor{cyan}{\noindent}}
\newcommand{\imran}[1]{\textcolor{brown}{\noindent}}
\fi
 
\newcommand{\mypar}[1]{\noindent\textbf{#1}}


\newcommand{\quickwordcount}[1]{%
  \immediate\write18{texcount -1 -sum -merge -q #1.tex output.bbl > #1-words.sum }%
  \input{#1-words.sum} words%
}

\title{Detecting natural disasters, damage, and incidents in the wild} 

\titlerunning{Detecting natural disasters, damage, and incidents in the wild}

%
\author{
Ethan Weber\inst{1} \and
Nuria Marzo \inst{1} \and
Dim P. Papadopoulos \inst{1} \and
Aritro Biswas \inst{1} \and \\
Agata Lapedriza \inst{1,3} \and
Ferda Ofli \inst{2} \and
Muhammad Imran \inst{2} \and
Antonio Torralba \inst{1}
}
\authorrunning{E. Weber et al.}
%
\institute{Massachusetts Institute of Technology 
\email{\{ejweber,nmarzo,dimpapa,abiswas,agata,torralba\}@mit.edu}
\and
 Qatar Computing Research Institute, HBKU  \\
 \email{\{fofli,mimran\}@hbku.edu.qa}
\\ \and
Universitat Oberta de Catalunya
}
\maketitle


\begin{abstract}

Responding to natural disasters, such as earthquakes, floods, and wildfires, is a laborious task performed by on-the-ground emergency responders and analysts. Social media has emerged as a low-latency data source to quickly understand disaster situations. While most studies on social media are limited to text, images offer more information for understanding disaster and incident scenes. However, no large-scale image datasets for incident detection exists. In this work, we present the Incidents Dataset, which contains 446,684 images annotated by humans that cover 43 incidents across a variety of scenes. We employ a baseline classification model that mitigates false-positive errors and we perform image filtering experiments on millions of social media images from Flickr and Twitter. Through these experiments, we show how the Incidents Dataset can be used to detect images with incidents in the wild. Code, data, and models are available online at \url{http://incidentsdataset.csail.mit.edu}.

\end{abstract}
\section{Introduction}

Rapid detection of sudden onset disasters such as earthquakes, flash floods, and other emergencies such as road accidents is extremely important for response organizations. However, acquiring information in the occurrence of emergencies is labor-intensive and costly as it often requires manual data processing and expert assessment. To alleviate these manual efforts, there have been attempts to apply computer vision techniques on satellite imagery, synthetic aperture radar, and other remote sensing data~\cite{gamba_rapid_2007,plank2014rapid,skakun2014flood,chehata2014object}.
Unfortunately, these approaches are still costly to deploy and they are not robust enough to obtain relevant data under time-critical situations. Moreover, satellite imagery is susceptible to noise such as clouds and smoke (i.e., common scenes during hurricanes and wildfires), and only provides an overhead view of the disaster-hit area.

On the other hand, studies show that social media posts in the form of text messages, images, and videos are available moments after a disaster strikes and contain information pertinent to disaster response such as reports of damages to infrastructure, urgent needs of affected people, among others~\cite{castillo2016big,imran2015processing}. However, unlike other data sources (e.g., satellite), social media imagery remains unexplored, mainly because of two important challenges. First, image streams on social media are very noisy, and disasters are not an exception. Even after performing a text-based filter, a large percentage of images in social media are not relevant to specific disaster categories. Second, deep learning models, that are the standard techniques used for image classification, are data-hungry, and yet no large-scale ground-level image dataset exists today to build robust computational models. 

In this work we address these challenges and investigate how to detect natural disasters, damage, and incidents in images. Concretely, our paper has the following three main contributions. First, we present the large-scale Incidents Dataset, which consists of 446,684 scene-centric images annotated by humans as positive for natural disasters (class-positives), types of damage or specific events that can require human attention or assistance, like traffic jams or car accidents. We use the term \emph{incidents} to refer to the 43 categories covered by our dataset (Sec.~\ref{section_dataset}). The dataset also contains an additional set of 697,464 images annotated by humans as negatives for specific incident categories (class-negatives).
As discussed in Sec.~\ref{sec_related_work}, the Incidents Dataset is significantly larger, more complete, and much more diverse than any other dataset related to incident detection in scene-centric images. 
Second, using the full set of 1.1M images in our dataset, we train different deep learning models for incident classification and incident detection. In particular, we use a slightly modified binary cross-entropy loss function, which we refer to as class-negative loss, that exploits our class-negative images. Our experiments in Sec.~\ref{section_experiments} show the importance of using class-negatives in order to train a model that is robust enough to be deployed for incident detection in real scenarios, where the number of negatives is large. Third, we perform extended incident detection experiments on large-scale social media image collections, using millions of images from Flickr and Twitter. These experiments, presented in Sec.~\ref{section_applications}, show how our model, trained with the Incidents Dataset and the class-negative loss, can be effectively deployed in real situations to identify incidents in social media images. We hope that the release of the Incidents Dataset will spur more work in computer vision for humanitarian purposes, specifically natural disaster and incident analysis.

\section{Related Work}
\label{sec_related_work}

\noindent{\textbf{Computer vision for social good.}}
Existing vision-based technologies are short of reaching out to diverse geographies and communities due to biases in the commonly used datasets. For instance, state-of-the-art object recognition models perform poorly on images of household items found in low-income countries~\cite{vries_does_2019}. To remedy this shortcoming, the community has made recent progress in areas including 
agriculture~\cite{efremova_aibased_2019,pryzant_monitoring_2017,rustowicz_semantic_2019,kaneko_deep_2019}, 
sustainable development~\cite{jean_combining_2016,helber_mapping_2019,workman2017unified}, 
poverty mapping~\cite{piaggesi_predicting_2019,watmough_socioecologically_2019,tingzon_mapping_2019}, 
human displacement~\cite{kalliatakis_displacenet_2019,kalliatakis_exploring_2019}, 
social welfare~\cite{bonafilia_building_2019,gebru_using_2017,gebru2017fine,nachmany_detecting_2019}, 
health~\cite{rehman_deep_2019,mckinney_international_2020,wu_deep_2019}, 
urban analysis~\cite{arietta2014city,can2018ambiance,kataoka2019ten,naik_streetscore_2014,naik_streetchange_2017,zhou2014recognizing}, 
and environment~\cite{kellenberger_when_2019,schmidt_visualizing_2019}. These studies, among many others, have shown the potential of computer vision to create impact for social good at a global scale.

\begin{figure}[t]
\centering
\includegraphics[width=\textwidth]{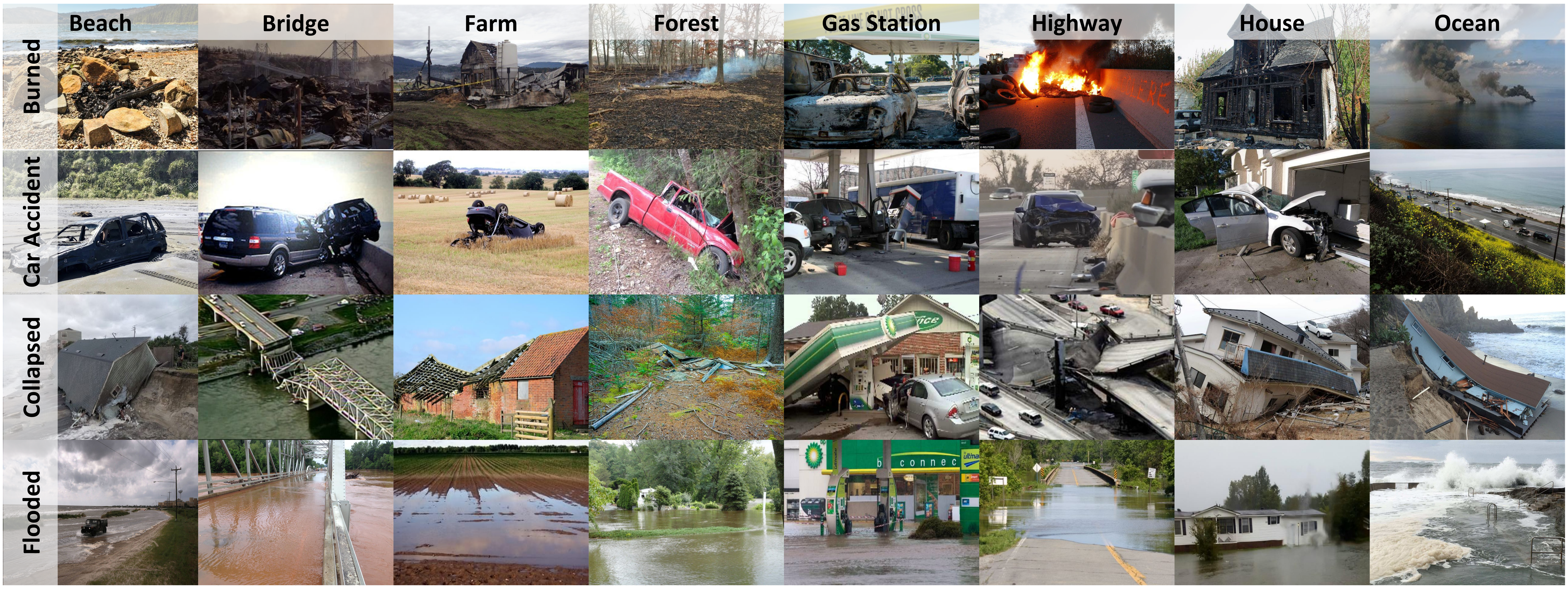}
\caption{\textbf{Example images from the Incidents Dataset.} Incidents (left) happen in many places (top), which we capture by having 43 incident and 49 place categories. Notice that a car accident can occur on a beach, farm, highway, etc. The place categories help by adding diversity to the dataset.}\label{fig:disaster_place_matrix}
\end{figure}

\noindent{\textbf{Incident detection on satellite imagery.}} 
There are numerous studies that combine traditional machine learning with a limited amount of airborne or satellite imagery collected over disaster zones
~\cite{turker_detection_2004,gamba_rapid_2007,chehata2014object,skakun2014flood,radhika2015cyclone,FernandezGalarreta:2015dn}. For a detailed survey, see~\cite{joyce2009review,dell_remote_2012,dong_comprehensive_2013,plank2014rapid}. Oftentimes, these studies are constrained to particular disaster events.
Recently, deep learning-based techniques have been applied on larger collections of remote-sensed data to assess structural damage~\cite{gueguen2015large,attari2016nazr,doshi_from_2018,li2019building,xu_building_2019,gupta_creating_2019} incurred by floods~\cite{nogueira2018exploiting,ben-haim_inundation_2019,rudner_multi3net_2019}, hurricanes~\cite{li2018semisupervised,sublime2019automatic}, and fires~\cite{radke2019firecast,doshi_firenet_2019}, among others. 
Some studies have also applied transfer learning~\cite{seo_revisiting_2019} and few-shot learning~\cite{oh_explainable_2019} to deal with unseen situations in emergent disasters.

\noindent{\textbf{Incident detection on social media.}}
More recently, social media has emerged as an alternative data source for rapid disaster response. Most studies have focused heavily on text messages for extracting crisis-related information~\cite{imran2015processing,reuter2018fifteen}. On the contrary, there are only a few studies using social media images for disaster response~\cite{peters2015investigating,chen2013understanding,daly2016mining,nguyen2017damage,nia2017building,alam2019processing,li2019identifying,pouyanfar2019unconstrained,Abavisani_2020_CVPR}. For example, \cite{nguyen2017damage} classifies images into three damage categories whereas \cite{nia2017building} regresses continuous values indicating the level of destruction. Recently, \cite{alam2019processing} presented a system with duplicate removal, relevancy filtering, and damage assessment for analyzing social media images. \cite{li2019identifying,pouyanfar2019unconstrained} investigated adversarial networks to cope with data scarcity during an emergent disaster event. 

\noindent{\textbf{Incident detection datasets.}}
Most of the aforementioned studies use small datasets covering just a few disaster categories, which limits the possibility of creating methods for automatic incident detection. In addition, the reported results are usually not comparable due to lack of public benchmark datasets, whether it be from social media or satellites~\cite{disaster_detection_survey}. One exception is the xBD dataset~\cite{gupta2019xbd}, which contains 23,000 images annotated for building damage but covers only six disasters types (earthquake, tsunami, flood, volcanic eruption, wildfire, and wind). On the other hand, \cite{gueguen2015large} has many more images but their dataset is constructed for detecting damage as anomaly using pre- and post-disaster images. There are also datasets combining social media and satellite imagery for understanding flood scenes~\cite{mediaeval_2017,mediaeval_2018} but they have up to 11,000 images only. In summary, existing incident datasets are small, both in number of images and categories. In particular, incident datasets are far, in size, from the current large datasets on image classification, like ImageNet~\cite{imagenet_2009} or Places~\cite{places_dataset}, which contain millions of labeled images. Unfortunately, neither ImageNet nor Places covers incident categories. Our dataset is significantly larger, more complete, and much more diverse than any other available dataset related to incident detection, enabling the training of robust models able to detect incidents in the wild.
\section{Incidents Dataset}
\label{section_dataset}

In this section, we present the Incidents Dataset collected to train models for automatic detection of disasters, damage, and incidents in scene-centric images.

\mypar{Incidents taxonomy.}
We create a fine-grained vocabulary of $233$ categories, covering high-level categories such as general types of damage (e.g., destroyed, blocked, collapsed), natural disasters including weather-related (e.g., heat wave, snow storm, blizzard, hurricane), water-related (e.g., coastal flood, flash flood, storm surge), fire-related (e.g., fire, wildfire, fire whirl), as well as geological (e.g., earthquake, landslide, mudslide, mudflow, volcanic eruption) events, and transportation and industrial accidents (e.g., train accident, car accident, oil spill, nuclear explosion). We then manually prune this extensive vocabulary by discarding categories that are hard to recognize from images (e.g., heat wave, infestation, famine) or by combining categories that are visually similar (e.g., snow storm and blizzard, or mudslide and mudflow). As a result of this pruning step, we obtain a final set of $43$ incident categories. 

\mypar{Image downloading and duplicate removal.}
Images are download from Google Images using a set of queries. To generate the queries and promote diversity on the data, we combine the $43$ incident categories with place categories. For the place categories, we select the 118 outdoor categories of Places dataset~\cite{places_dataset} and merge categories belonging to the same super-category (e.g., topiary garden, Japanese garden, vegetable garden are merged into garden). After this process we obtain $49$ different place categories. By combining incident and place categories, we obtain a total of $43 \textrm{ incidents} \times 49 \textrm{ places} = 2107$ pairs. Each pair is extended with incidents and places synonyms to create queries such as ``car accident in highway" and ``car wreck in flyover'', or ``blizzard in street'' and ``snow storm in alley.'' We obtain 10,188 queries in total and we download all images returned from Google Images for each query, resulting in a large collection of 6,178,192 images. After that, we perform duplicate image removal as follows: we extract feature vectors from each image with a ResNet-18~\cite{he2016deep} model trained on Places~\cite{places_dataset} and we cluster duplicate images with a radius-based Nearest Neighbor algorithm. This results in 3,487,339 unique images. 

\mypar{Image labeling.}
Images obtained through Google Images are noisy, and they may not necessarily be relevant to the query they are downloaded for. Rather, the results may contain non-incident images with similar appearances (e.g., airplanes but not airplane accidents, fireplaces but not dangerous fires, bicycles and not bicycle accidents, etc.), images with other incidents, or completely random images. 
To clean the data we ask annotators to manually verify the images using the Amazon Mechanical Turk (MTurk) platform.
Workers are shown a batch of images, and they have to answer whether each image belongs to a specific category or not. In particular, the interface used for image annotation is similar to \cite{places_dataset}. Each image is annotated by one annotator. Each annotation batch contains 100 images, including 15 control images (10 positives and 5 negatives). Annotation batches are accepted when the accuracy in the control images is above $85\%$. Otherwise the annotations of the batch are discarded.

\begin{figure}[t]
\centering
\includegraphics[width=1.0\textwidth]{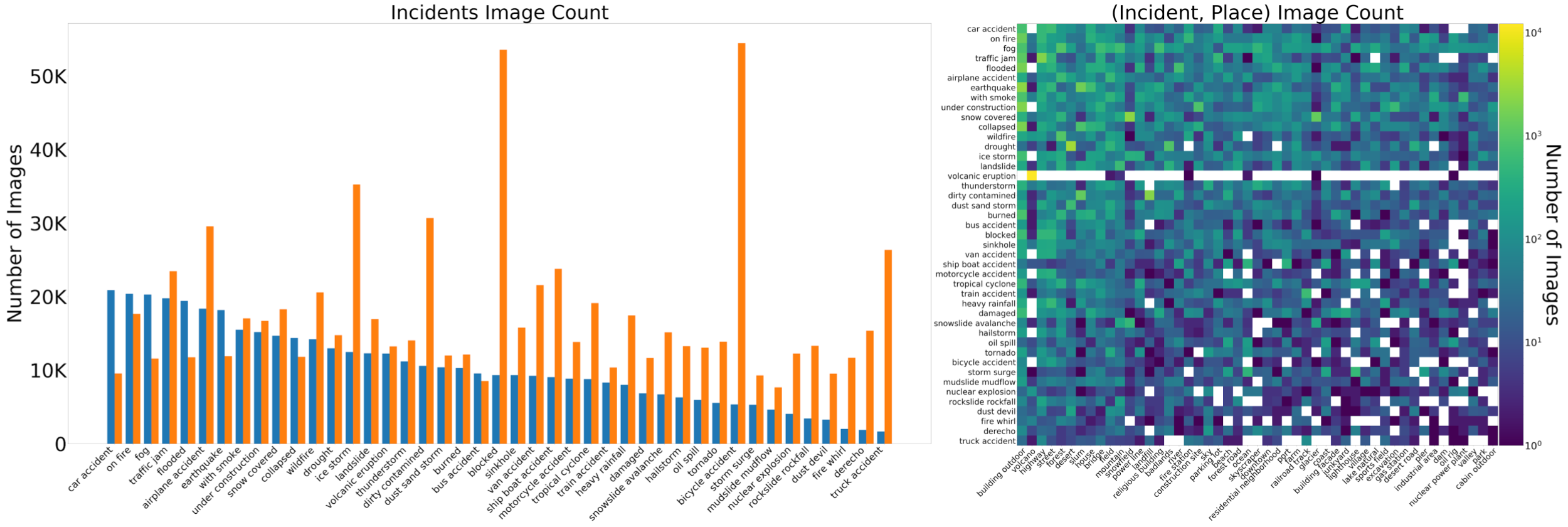}
\caption{\textbf{Dataset composition.} The number of positive (blue) and negative (orange) labeled incident images is shown on the left and the distribution of images for (incident, place) combinations is shown on the right. The dataset contains incidents in many different scenes. White cells indicate the \emph{unlikely} (incident, place) combinations for which the Incidents Dataset does not contain any images (e.g., ``car accident in volcano'').}
\label{fig:dataset_composition}
\end{figure}

The images are annotated in several stages. First, we label 798,316 images from the initial 3,487,339 image collection, using the queries the images are downloaded from. For example, the images downloaded with the query ``car accident in village'' are labeled as positives or negatives for the class ``car accident.'' This results in 193,648 class-positive incident images and 604,668 class-negatives. Class-negative images are those that we know do not show a specific incident class but they may contain another incident category. 
After the first annotation stage, we train a temporary incident recognition model, as described in Sec.~\ref{section_model}, to determine which images to label next. We send images whose incident category confidence scores were greater than $0.5$ to MTurk to get more class-positive and class-negative labels. This process is repeated until obtaining 446,684 positive incident images. Finally, these 446,684 images are sent to MTurk for annotation on place categories using the same interface. In this case, each image is assessed for the place category of its original query (e.g., an image downloaded with the query ``wildfire in forest" is labeled as positive or negative for the ``forest'' category). Eventually, we obtain 167,999 images with positive place labels. The remaining images have negative place labels.

\mypar{Dataset statistics.} The Incidents Dataset contains 1,144,148 labeled images in total. Of these, 446,684 are class-positive incident images, 167,999 of which have also positive place labels. Fig.~\ref{fig:disaster_place_matrix} shows some sample images from our dataset. Fig.~\ref{fig:dataset_composition} shows the number of images per incident, place, and combined (incident, place). Although the common practice when collecting datasets is just to keep images with positive labels, we will show in Sec.~\ref{section_model} and Sec.~\ref{section_experiments} that class-negative images are particularly valuable for incident detection in the wild because they can be used as hard negatives for training.

\section{Incident model}\label{section_model}

In this section, we present our model for recognizing and detecting different incident types in scene-centric images. 

\mypar{Multi-task architecture.}
The images in our Incidents Dataset are accompanied with an incident and a place label (see Sec.~\ref{section_dataset}). We choose to build a single model that jointly recognizes incident and place categories following a standard multi-task learning paradigm~\cite{caruana1997multitask,rebuffi2017learning,simonyan2014two}. This architecture offers efficiency as it can jointly recognize incidents and places, and we also did not observe any difference in the performance when training a model for a single task. In our experiments, we employ a Convolutional Neural Network (CNN) architecture with two task-specific output layers. Specifically, our network is composed of a sequence of shared convolutional layers followed by two parallel branches of fully-connected layers, corresponding to incident and place recognition tasks.

\mypar{Training with a cross-entropy loss.}
The standard and most successful strategy for training an image classification model (either for incidents or places) is to employ a cross-entropy loss on top of a softmax activation function for both outputs of the network. Note that this is the standard procedure for single-label classification of objects~\cite{imagenet_2009}, scenes~\cite{places_dataset} or actions~\cite{simonyan2014two}.

In our real-world scenario of detecting incidents in social media images, many of the test scene-centric images do not belong to any of the incidents categories and they should be classified as images with ``no incident.'' This can be handled by adding an extra neuron in the output layer that should fire on ``no incident'' images. Notice that this requires training the model with additional absolute negative images, i.e., images that do not show any incident.

\mypar{Training with a class-negative loss.}
Even during an incident, the number of images depicting the incident is only a small proportion of all the images shared in social media. For this reason, our task of finding incidents in social media imagery is more closely related to that of detection \cite{Girshick_2014_CVPR,Lin_2017_ICCV,Shrivastava_2016_CVPR} than classification. In particular, our model must find positive examples out of a pool of many challenging negatives (e.g. a chimney with smoke or a fireplace are not disaster situations, yet they share visual features similar to our ``with smoke" and ``on fire" incident categories). To handle this problem and mitigate false positive detections, either the training process can be improved \cite{Durand_2019_CVPR,lee2018training} or the predictions can be adjusted at test time \cite{liang2017enhancing,leeNIPS2018}. For our task, we choose to modify our training process to incorporate class-negatives.

In particular, similar to \cite{Durand_2019_CVPR}, we modify a binary cross entropy (BCE) loss to use partial labels for single-label predictions. Our partial labels consist of both the class-positive and class-negative labels obtained during the image annotation process (Sec.\ref{section_dataset}). Notice that class-negative images are, in fact, hard negatives for the corresponding classes because of the way they were selected during labeling: they are either false-positive results returned from the image search engine or false-positive predictions with high confidence scores using our model. More formally, we modify BCE by introducing a weight vector to mask the loss where we've obtained partial labels. This is given by the equation:

\begin{align}
    \textrm{Loss} &= \sum_{x_{i}, y_{i}, w_{i} \in X, Y, W}\left[w_{i}[y_{i}\log(A(x_{i})) + (1 - y_{i})\log(1-A(x_{i}))]\right]\label{eq:loss}
\end{align}

\noindent where $A$ is the activation function (typically a sigmoid), $X$ the prediction, $Y$ the target, and $W$ the weight vector. $X,Y,W \in \mathbb{R}^{N}$, and $N$ is the number of classes.

For a training image with a class-positive label, we set $y_{i} = 1$ and $W = 1^{N}$ because we can conclude all information is known (i.e., due to our single-label assumption, the image is considered as negative for all the other classes). For a class-negative training image of the class $i$, we set $y_{i} = 0$ and $w_{i} = 1$. We do not set $W = 1^{N}$ 
in this case because we do not have ground truth positive or negative labels for the rest of the classes (different incidents may or may not appear in the image). Hence, for any unknown class $j$, i.e., $j \ne i$, we set $w_{j} = 0$.

The final loss $\mathcal{L}$ is given by the sum of the incidents loss $\mathcal{L}_{d}$ and the place loss $\mathcal{L}_{p}$, where both $\mathcal{L}_{d}$ and $\mathcal{L}_{p}$ are given by Eq.~(\ref{eq:loss}).
\section{Experiments on the Incidents Dataset}\label{section_experiments}

\subsubsection{Data.} 
We split the images of the Incidents Dataset into training ($90\%$), validation ($5\%$), and a test ($5\%$). As a reference, the training set contains 1,029,726 images, with 401,875 class-positive and 683,572 class-negative incident labels, and 151,665 class-positive and 265,415 class-negative place labels. Note that an image may have more than one class-negative label. Since the number of class-positive place labels is much lower than the number of class-positive incident labels, we augment the training set with 42,318 images from the Places dataset~\cite{places_dataset}. However, while training, we do not back-propagate the incidents loss on the additional Places images (which have no incident) since we already have class-negatives for the incidents that are better negative examples than these images from Places with no incidents.

The \emph{test set} contains 57,215 images, and we 
also construct an \emph{augmented test set} that is enriched with 2,365 extra images from Places that we assume contain no incidents. Unlike other image classification datasets, our \emph{test set} contains class-negative images, which are important to evaluate the ability of a model to detect incidents in test images.

\mypar{Incident classification.}
We first evaluate the ability of our model to classify an image to the correct incident category, using just the images from the test set that belong to an incident category.
Note that this experiment is similar to a within-the-dataset classification task where every test image belongs to a target category.
We use a ResNet-18~\cite{he2016deep} as backbone and train the model using the class-negative loss. We evaluate the incidents classification accuracy only on the part of test set that has positive incident labels. The top-1 accuracy is 77.3\%, while the top-5 accuracy is 95.9\%.  As a reference, the performance of the same architecture trained on the same images but with a cross-entropy loss, which is a more standard choice for this classification task, is only slightly better, with 78.9\% top-1 and 96.3\% top-5 accuracy. 

\begin{table}[t]
\centering
\caption{\textbf{Ablation study.} Performance comparison of the proposed model under different settings on both test sets. The best mAP is achieved by the model that uses CN loss with additional Places images as well as class negatives.}
\label{table:map_table}
\resizebox{\textwidth}{!}{%
\begin{tabular}{c c c c c c c c}
\toprule
              &       \multicolumn{3}{c}{Training with}                     & \multicolumn{2}{c}{Test Set} & \multicolumn{2}{c}{Augmented Test Set}\\ \cmidrule(lr){2-4} \cmidrule(lr){5-6} \cmidrule(lr){7-8}
Architecture  & Loss & Class Negatives & Additional Places Images & Incident mAP        & Place mAP          & Incident mAP                 & Place mAP\\
\cmidrule(lr){1-1}\cmidrule{2-2}\cmidrule(lr){3-3}\cmidrule(lr){4-4}\cmidrule(lr){5-5}\cmidrule(lr){6-6}\cmidrule(lr){7-7}\cmidrule(lr){8-8}
ResNet-18 & CE         &  &\checkmark& 62.04 & \textbf{47.85}& 60.60 & 53.60 \\
ResNet-18  & CN         &  &\checkmark& 61.15 & 46.61 & 59.88 & 53.41 \\
ResNet-18  & CN & \checkmark          &  & 66.59 & 46.59 & 65.39 & 52.82\\
ResNet-18  & CN & \checkmark & \checkmark  &  66.35 & 46.71& 65.76 & 62.04\\
ResNet-50 & CN  & \checkmark & \checkmark & \textbf{67.65} & \textbf{47.56}&\textbf{67.19}&\textbf{63.20}\\
\bottomrule
\end{tabular}%
}
\end{table}

\mypar{Incident detection.}
We consider here a more realistic scenario of detecting incidents in images, evaluating the performance of the model on the whole test set that also includes images with negative labels. We measure this performance using the average precision (AP) metric, and we report the mean over all categories (mAP) for both the incidents and the places.  

The obtained results are shown in Tab.~\ref{table:map_table}, that presents, in fact, an ablation study exploring the use of different model architectures (ResNet-18 and ResNet-50), losses (cross-entropy and class-negative), and training data. Each model is pre-trained on the Places365 dataset~\cite{places_dataset} for the task of scene classification and then fine-tuned with the corresponding Incidents training data until convergence (at least 10 epochs). We used the Adam optimizer with an initial learning rate of 1e-4 and a batch size of 256 images, with shuffling. For each model, we report the incident and the place mAP on both the \emph{test set} and the \emph{augmented test set}.

\begin{figure}[t]
\centering
\includegraphics[width=\textwidth]{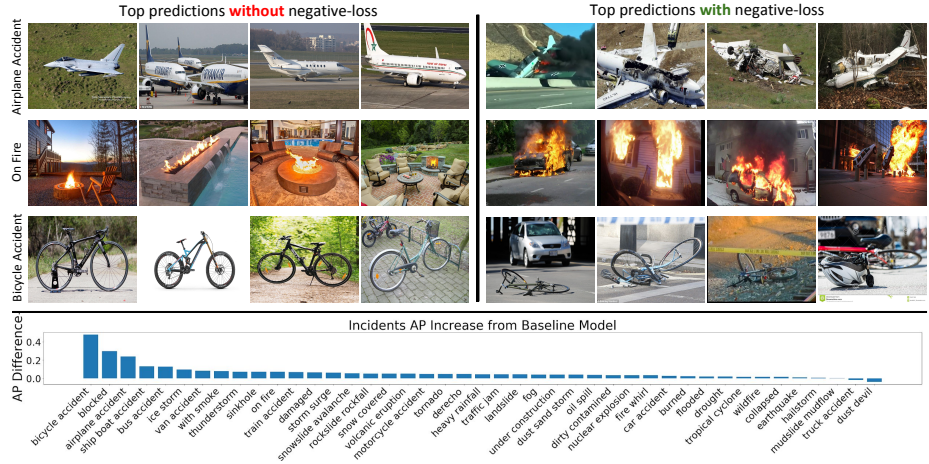}
\caption{\textbf{Using the class-negative loss.} Top confidence images for ``airplane accident'', ``on fire'', and ``bicycle accident'' categories when training without (left) and with (right) the class-negative loss. (Bottom) We report incident AP increments achieved by our model over the baseline model.}
\label{fig:With_Without_Negatives}
\end{figure}

We observe that the incident mAP significantly improves by 4.3\% (on the test set) to 5.2\% (on the augmented test set) when we move from the cross-entropy (CE) loss to the class-negative loss (CN) using the class negatives (first and fourth row of Tab.~\ref{table:map_table}).  
Fig.~\ref{fig:With_Without_Negatives} shows some top-ranked images for three incident categories by these two models. We can observe that, without using the class negatives during training, the model is not able to distinguish the difference between a fireplace and a house on fire or detect when a bicycle is broken because of an accident. The bottom of Fig.~\ref{fig:With_Without_Negatives} shows 
the change in AP per incident category achieved by the CN model over the CE model.
Notice that for nearly all incident categories the AP is much higher with CN model.

As a reference, the performance of the CN loss without using any class negatives, which corresponds to the standard BCE loss, is only less than 1\% worse than the CE (first and second row of Tab.~\ref{table:map_table}). Using additional Places images during training does not affect the incident detection but it vastly improves the place detection, especially in the case of the augmented test set, where mAP increases by 9.2\% (third and fourth row of Tab.~\ref{table:map_table}). Switching from a ResNet-18 to a deeper ResNet-50 architecture gives an extra final boost of incident mAP performance by 1.3\% (fifth row of Tab.~\ref{table:map_table}).

To further demonstrate the improved performance of our model trained with the CN loss (final), we compare it against the model trained with a CE loss (baseline) on 208 hand-selected hard-negative images used for MTurk quality control and not seen during training. Our final model recognizes 176 images correctly as true negatives with confidence score below 0.5 ($85\%$ accuracy) while the baseline model predicts the majority of them as false positives ($30\%$ accuracy).

\begin{figure}[t]
\centering
\includegraphics[width=\textwidth]{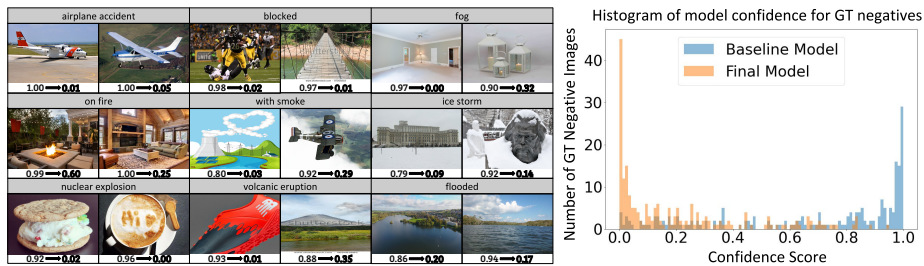}
\caption{\textbf{GT negative test.} (Left) Sample images withheld for quality control on MTurk are GT negatives not seen by the model during training. We report the changes in confidence scores between the baseline and final model below each image. (Right) We visualize the distribution of confidence scores obtained by both models for all 208 GT negative images. Our final model is more conservative when predicting incident confidence scores for hard-negative examples.}
\label{fig:ground_truth_negatives_test}
\end{figure}

In Fig.~\ref{fig:ground_truth_negatives_test}, we investigate the changes in the confidence scores between the baseline and final models. Fig.~\ref{fig:ground_truth_negatives_test}(left) displays some qualitative examples of false positives for different incident categories. 
We observe that the confidence scores significantly decrease when using the final model. More concretely, the final model does not associate airplane features blindly to airplane accident, does not confuse rivers with flood scenes, or does not mistake clouds as smoke. Fig.~\ref{fig:ground_truth_negatives_test}(right) shows the distribution of confidence scores obtained by the baseline and final models. Notice that a perfect detector should assign 0 score to all of these images. 
Overall, this analysis shows, consistently with the other experiments explained in this section, how our final model is more robust against difficult cases, which is very important for filtering disaster images in the wild.
\section{Detecting Incidents in Social Media Images}\label{section_applications}

\begin{figure}[t]
\centering
\includegraphics[width=\textwidth]{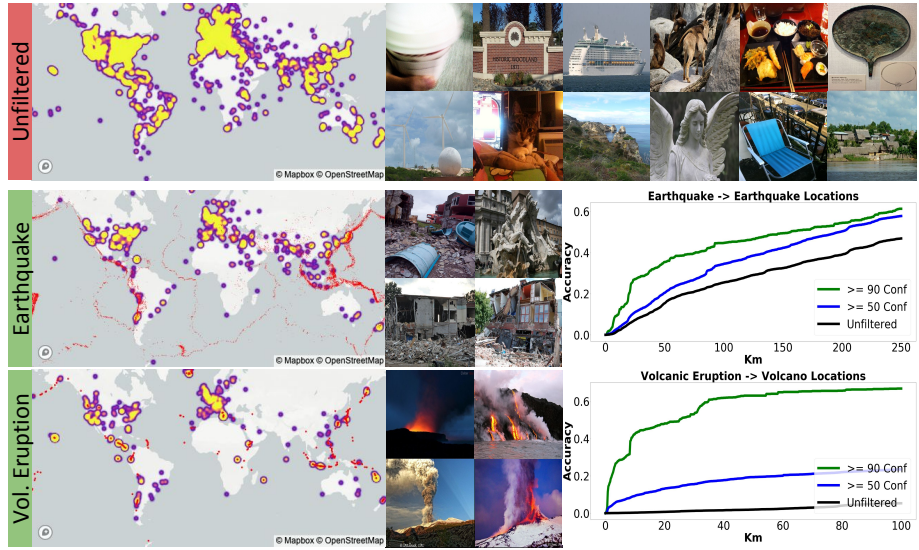}
\caption{\textbf{Filtering Flickr images.} (Top) Left: map visualization of Flickr image locations (complete unfiltered set). Right: random Flickr images. (Middle) Earthquake filtering. Left: map visualization of the location of images filtered by the earthquake category (earthquake epicenters are displayed as red dots). Middle: examples of images filtered with the earthquake category. Right: Accuracy@XKm, defined as the percent of images within X kilometers from an epicenter. When filtering with a confidence threshold above $0.9$ (green), images are much closer to epicenters than in the unfiltered case (black). (Bottom) Volcanic eruptions and volcanoes, with the same structure as the earthquake experiment.}
\label{fig:flickr_filtering}
\end{figure}

In this section, we examine how our incident detection model, trained with class-negative loss, performs in three different real-world scenarios using millions of images collected from two popular social media platforms: Flickr and Twitter. 

\begin{figure}[t]
\centering
\includegraphics[width=\textwidth]{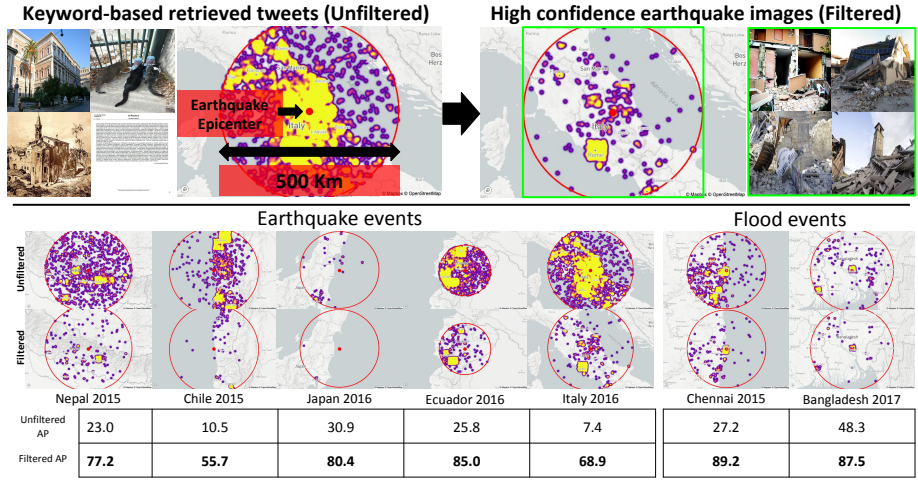}
\caption{\textbf{Twitter image filtering.} (Top) Experiment outline for the earthquake filtering (we follow the same outline for floods): we consider all tweets within a 250 Km radius of the epicenter of a specific event and then we filter the images for the earthquake category. Left part shows image examples and location of the unfiltered images, while right part shows locations and examples of filtered images. 
(Middle) Locations of unfiltered (top) and filtered images (bottom) are shown for each one of the seven events (five earthquakes and two floods), respectively.
(Bottom) Ground-truth labels obtained from MTurk for each event are used to compute the AP for unfiltered (top row) and filtered (bottom row) images. Notice that our model significantly outperforms the unfiltered baseline.}
\label{fig:disaster_image_filter}
\end{figure}

\subsection{Incident detection from Flickr images}
\label{flickr}

The goal of this experiment is to illustrate how our model can be used to detect specific incident categories in the wild. For this purpose, we use 40 million geo-tagged Flickr images obtained from the YFCC100M dataset~\cite{thomee2016yfcc100m}. Since the images have precise geo-coordinates from EXIF data, we can use our incident detection model to filter for specific incidents and compare distance to ground-truth locations. We evaluate only earthquake and volcanic eruption incidents in this experiment as we could find reasonable ground-truth data to compare the results. Specifically, we downloaded the GPS coordinates, i.e., latitude and longitude, of volcanoes from the National Oceanic and Atmospheric Administration (NOAA) website\footnote{\url{https://www.noaa.gov/}} and a public compilation of earthquake epicenters\footnote{\url{https://raw.githubusercontent.com/plotly/datasets/master/earthquakes-23k.csv}}. We employ an Accuracy@XKm metric~\cite{accuracy_x_km} to determine whether the predicted incident is correct or not. More concretely, we compute the percentage of images within X Km from the closest earthquake epicenter or volcano, respectively. We randomly sample images and report metrics for (i) unfiltered images, (ii) images with model confidence above $0.5$, and (iii) images with model confidence above $0.9$. Fig.~\ref{fig:flickr_filtering} shows that detected earthquake and volcanic eruption incidents appear much closer to expected locations when compared to random images.

\subsection{Incident detection from Twitter images}
\label{twitter_filtering}

In this experiment we aim to detect earthquakes and floods in noisy Twitter data posted during actual disaster events. We collected data from five earthquake and two flood events using event-specific hashtags and keywords. In total, 901,127 images were downloaded. Twitter GPS coordinates are not nearly as precise as the Flickr ones, so we consider only the 39,494 geo-located images within 250 Km from either (i) the earthquake epicenter or (ii) the flooded city center.

For all seven events shown in Fig.~\ref{fig:disaster_image_filter}, we use MTurk to obtain ground-truth human labels (i.e., earthquake or not, and flood or not) for images within the considered radius. 
Then, we compare the quality of the initial set of the keyword-based retrieved Twitter images (unfiltered) to the quality of images retained by our model (filtered). 
We report the average precision (AP) per event 
for both earthquakes and floods. 
When considering all earthquake events and flood events, we obtain a average AP of 73.9\% and 89.1\% compared to 
the baseline AP of 11.9\% and 28.2\%, respectively. The baseline AP is the AP averaged over multiple trials of randomly shuffling the images, and it is given as a reference.

\begin{figure}[t]
\centering
\includegraphics[width=\textwidth]{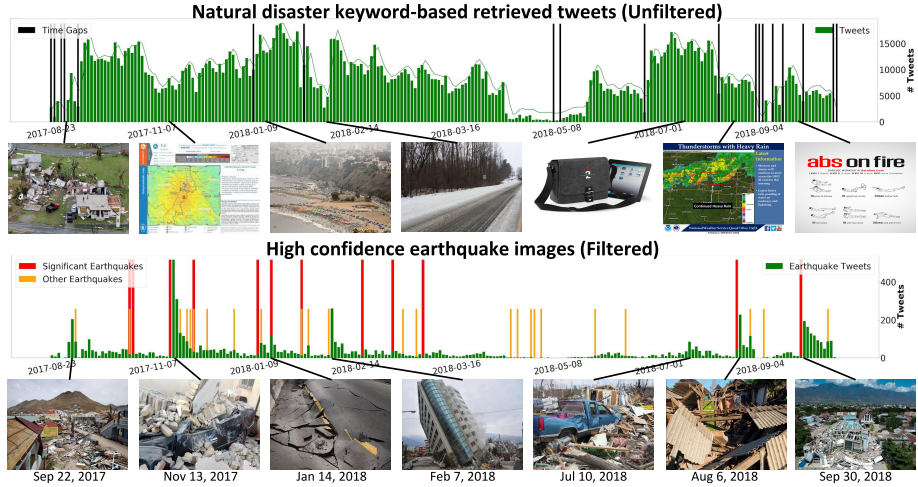}
\caption{\textbf{Finding peaks in earthquake tweets.} (Top) Histogram of tweets obtained from Twitter using natural disaster keywords from 2017-2018. Black lines indicate periods of time when our data collection server was inactive. (Bottom) Number of tweets with earthquake images per day after filtering with at least $0.5$ confidence. For significant earthquakes (above 6.5 magnitude), we notice an increase in earthquake images immediately after the event. Furthermore, we notice a spike on July 20, 2018 not reported in the NOAA database. We manually checked the tweets and found images referring to a severe flood in Japan, indicating that the flood damage may resemble earthquake damage.}
\label{fig:temporal_analysis}
\end{figure}

\subsection{Temporal monitoring of incidents on Twitter}\label{twitter_temporal}

In this section we demonstrate how our model can be used on Twitter data stream to detect specific incidents in time. 
To test this, we downloaded 1,946,850 images from tweets containing natural disaster keywords (e.g., blizzard, tornado, hurricane, earthquake, active volcano, coastal flood, wildfire, landslide) from Aug. 23, 2017 to Oct. 15, 2018. To quantify detection results, we obtained ground-truth event records from the ``Significant Earthquake Database", the ``Significant Volcanic Eruption Database", and the ``Storm Events Database" of NOAA. The earthquake and volcanic eruptions ground-truth events are rare \textit{global} events, while the storms (floods, tornadoes, snowstorms and wildfires) are much more frequent but reported only for the \textit{United States}. We filter images with at least $0.5$ confidence and compare against the databases (Fig.~\ref{fig:temporal_analysis}).

\begin{figure}[t]
\centering
\includegraphics[width=\textwidth]{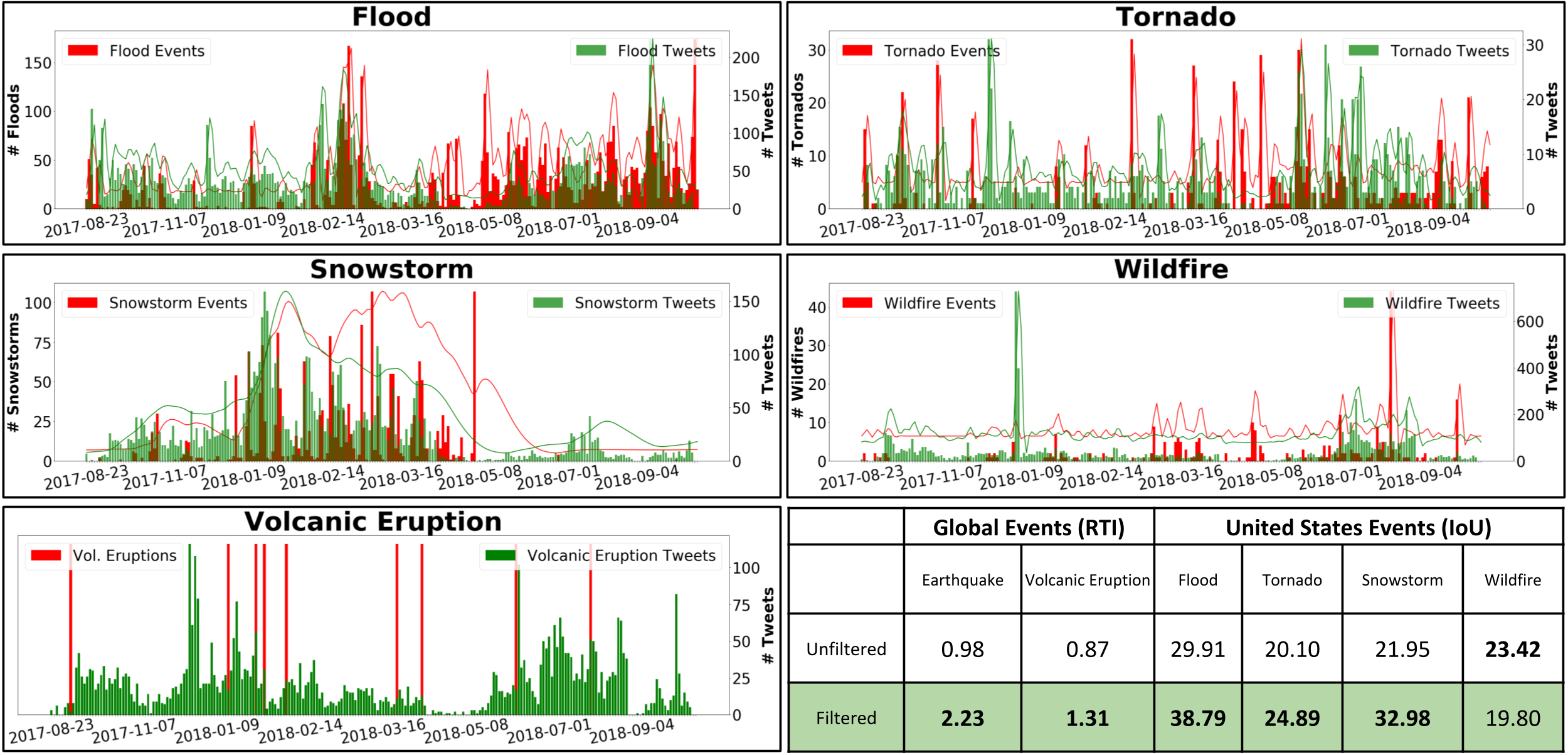}
\caption{\textbf{Temporal tweet filtering results.} (Rows 1-2) For frequent events in the United States, we filter tweets for floods, tornadoes, snowstorms, and wildfires images and compare with ground truth frequency events obtained from NOAA. (Bottom Left) Filtered volcanic eruption images with ground truth events. (Bottom Right) Reported mRTI for global events and IoU for common US events.}
\label{fig:temporal_analysis_all}
\end{figure}

For earthquakes and volcanic eruptions, we report average Relative Tweet Increase (RTI) inspired by~\cite{avvenuti2014ears}. 
$\textrm{RTI}_{e} = \sum_{d=e}^{e+w} N_{d}/\sum_{d=e}^{e-w} N_{d}$, where $N_{d}$ is the number of relevant images posted on day $d$, $e$ is the event day (e.g., day of earthquake), and $w$ is an interval of days. We use $w=7$ for our analysis to represent a week before and after an event. An $RTI$ of 2 means that the average number of tweets in the week following an event is twice as high as the average number the week before. After filtering, the mean RTI ($\textrm{mRTI} = \sum_{e \in E} RTI_{e}/|E|$) shows an average of $2.42$ folds increase in tweets the week after an earthquake and $1.31$ folds after a volcanic eruption (Fig.~\ref{fig:temporal_analysis_all}).

We notice that the mRTI would be even better if the ground truth databases were exhaustive. On Nov. 27, 2017 we detect the highest number of volcanic eruption images, but observe no significant eruption in the database. Looking into this, we found that Mount Agung erupted the same day, which caused the airport in Bali, Indonesia to close and left many tourists stranded\footnote{\url{https://en.wikipedia.org/wiki/Mount_Agung}}.

For the more common events (e.g., tornadoes and snowstorms), we measure the correlation between tweet frequency and event frequency. We normalize both histograms, smooth with a low-pass filter, and report intersection over union (IoU) for United States incidents in Fig.~\ref{fig:temporal_analysis_all}. We notice an increase in IoU after filtering for flood, tornado, and snowstorm images. For wildfires, we notice a decrease in IoU and attribute this to the large spike in tweets in December 2017. Frequency correlation does not represent damage extent. In fact, a destructive wildfire occurred in California on Dec. 4, 2017 burning 281,893 acres\footnote{\url{https://en.wikipedia.org/wiki/Thomas_Fire}}.
\section{Conclusion}
In this paper, we explored how to automatically and systematically detect disasters, damage, and incidents in social media images in the wild. We presented the large-scale Incidents Dataset, which consists of 446,684 human-labeled scene-centric images that cover a diverse set of $43$ incident categories (e.g., earthquake, wildfire, landslide, tornado, ice storm, car accident, nuclear explosion, etc.) in various scene contexts. Different from common practice, the Incidents Dataset includes an additional 697,464 class-negative images which can be used as hard negatives to train a robust model for detecting incidents in the wild.
To that end, we also used a class-negative loss that capitalizes on this phenomenon. We then showed how the resulting model can be used in different settings for identifying incidents in large collections of social media images. We hope that these contributions will motivate further research on detecting incidents in images, and also promote the development of automatic tools that can be used by humanitarian organizations and emergency response agencies.

\subsubsection*{Acknowledgments}

This work is supported by the CSAIL-QCRI collaboration project and RTI2018-095232-B-C22 grant from the Spanish Ministry of Science, Innovation and Universities.

\bibliographystyle{splncs04}
\bibliography{bibliography}
\end{document}